\documentclass[10pt,twocolumn,letterpaper]{article}

\usepackage{cvpr}
\usepackage{cite}
\usepackage{times}
\usepackage{epsfig}
\usepackage{subfig}
\usepackage{graphicx}
\usepackage{amsmath}
\usepackage{amssymb}
\usepackage{multirow}
\usepackage{diagbox}
\usepackage{bm}
\usepackage{authblk}

\cvprfinalcopy 

\def\cvprPaperID{6070} 

\ifcvprfinal\fi
\begin{document}

\title{Temporal Hockey Action Recognition via Pose and Optical Flows}

\makeatletter
\renewcommand\AB@affilsepx{, \protect\Affilfont}
\makeatother
\author[*]{Zixi Cai}
\author[+]{Helmut Neher}
\author[+]{Kanav Vats}
\author[+]{David Clausi}
\author[+]{John Zelek}
\affil[*]{Tsinghua University}
\affil[+]{University of Waterloo}

\maketitle
\ifcvprfinal\thispagestyle{empty}\fi

\begin{abstract}
    Recognizing actions in ice hockey using computer vision poses challenges due to bulky equipment and inadequate image quality.  A novel two-stream framework has been designed to improve action recognition accuracy for hockey using three main components.  First, pose is estimated via the Part Affinity Fields model to extract meaningful cues from the player.  Second, optical flow (using LiteFlownet) is used to extract temporal features.  Third, pose and optical flow streams are fused and passed to fully-connected layers to estimate the hockey player’s action. A novel publicly available dataset named HARPET (Hockey Action Recognition Pose Estimation, Temporal) was created, composed of sequences of annotated actions and pose of hockey players including their hockey sticks as an extension of human body pose.  Three contributions are recognized.  (1) The novel two-stream architecture achieves 85\% action recognition accuracy, with the inclusion of optical flows increasing accuracy by about 10\%.  (2) The unique localization of hand-held objects (e.g., hockey sticks) as part of pose increases accuracy by about 13\%.  (3) For pose estimation, a bigger and more general dataset, MSCOCO, is successfully used for transfer learning to a smaller and more specific dataset, HARPET, achieving a PCKh of 87\%.
\end{abstract}

\section{Introduction}

Vison-based human action recognition has gained increasing attention in the past few years because of broad applications in smart surveillance systems, smart elderly assistance, human-computer interaction, and sports as examples. Many challenges, such as lack of data, noisy data from bulky clothing and equipment, small human size due to camera position, similarities between foreground and background, and motion blur from high speed human actions, exist in many applications. One application that emulates these challenges is ice hockey.

\begin{figure*}
\begin{center}
\includegraphics[width=0.9\linewidth]{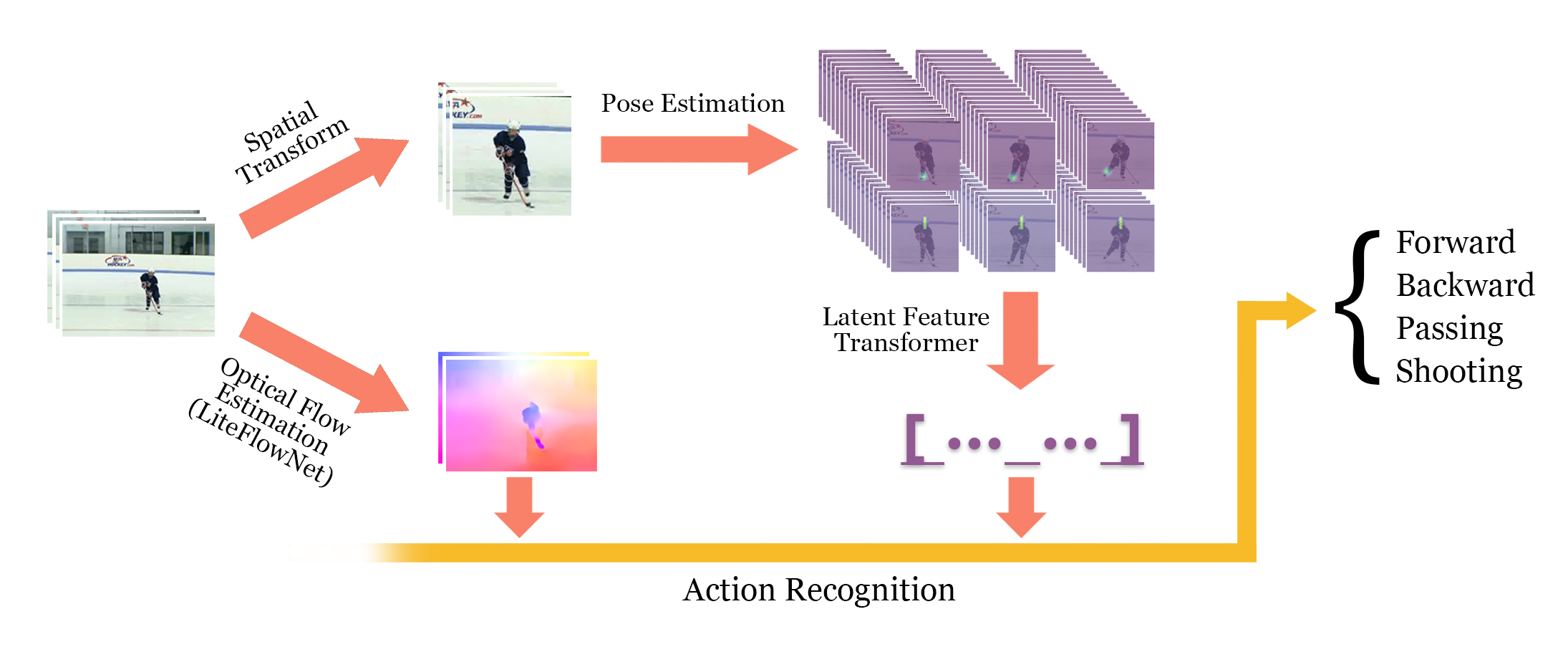}
\end{center}
   \caption{Overall pipeline. Our method takes sequence of 3 images as input. Part confidence maps and part affinity fields are predicted for each spatially transformed image, and converted into latent joint feature vector. Optical flows are generated in the second stream. Optical flows and the latent feature vector are used as the input of the action recognition component, which predicts probabilities of skating forward, skating backward, passing and shooting.}
\label{fig:overall_architecture}
\end{figure*}

 This paper focuses on incorporating pose information and optical flow for action recognition in a unified two-stream architecture (shown in Fig. \ref{fig:overall_architecture}) to provide high-level features unique to pose estimation and optical flow to depict motion, thus, improving the overall accuracy of action recognition. It also demonstrates the complementary nature of pose estimation and optical flow in improving action recognition accuracy. The two-stream architecture analyzes pose and temporal features via a convolutional neural network (CNN), then the outputs of the two streams are concatenated via fully-connected layers.

Although many works explore action recognition in videos on large benchmark datasets such as UCF101 and HMDB, few focus on sport videos~\cite{baccouche2010soccer, liu2015realtimesports, kong2008groupsoccer, karpathy2014largescale}. To date, there are no publicly available temporal action recognition datasets in hockey considering individual players; one dataset explores multiple players temporally ~\cite{sozykin2018imbalanced}, while, another dataset only considers still images with no temporal information considered ~\cite{fani2017hockeyar}. To solve this problem, a novel publicly available dataset, known as  HARPET (Hockey Action Recognition Pose Estimation Temporal), comprised of hockey image sequences (three images per sequence) captured by a single RGB camera are used, with annotations including pose (comprised of 18 joints including the hockey stick) and actions, is generated. Four types of actions are considered: skating forward, skating backward, passing and shooting. The dataset contains around 100 sequences per class, with around 1200 images in total.

Testing on HARPET dataset, the two-stream architecture obtains around 85\% end-to-end accuracy. For pose estimation model, due to small number of examples in dataset, transfer learning is leveraged to reduce overfitting and demonstrated to be effective with around 87\% PCKh@0.5~\cite{mpii}. It is also demonstrated that localization of hand-held objects can improve the accuracy of sports action recognition, which to the best of our knowledge, has not been explored in previous works.

The rest of the paper is organized as follows. In Section~\ref{section:background}, we view papers on action recognition, highlighting two-stream-based and pose-based frameworks, and discuss works on hockey action recognition. The architecture, comprising of pose estimation and action recognition models, is illustrated and implementation details are explained in Section~\ref{section:methodology}. We evaluate accuracy of both pose estimation and end-to-end action recognition on HARPET in Section~\ref{section:test}.



\section{Background}\label{section:background}

Action recognition is a widely researched topic which, before the advent of deep networks, employed hand-crafted features, dense trajectories~\cite{wang2011dt} and improved dense trajectories~\cite{wang2013idt}. State-of-the-art action recognition models incorporate these features in action recognition~\cite{wang2013dtmb, cheronICCV15, choutas2018potion, jain2013exploit, jhuang2013towards}. Recently, deep networks have shown promising action recognition accuracy through the use of 3D convolutions in demonstrating better capability of capturing spatiotemporal latent structure in videos than 2D convolutions~\cite{tran2018closer, ji2013conv3d, tran2015c3d, varol2018longterm, taylor2010convlearning, carreira2017i3d}. The major downside of 3D CNNs is the large number of parameters, making it easy to overfit on small datasets which is common in many practical applications. Also, the use of recurrent neural networks, which are manifested to be adept at modeling sequential data are explored~\cite{baccouche2010soccer, ng2015beyond, donahua2017lrcn}. To summarize, mainstream methods improve action recognition using several overlapping categories including: hand-crafted features~\cite{wang2011dt, wang2013idt, sun2018off}, two-stream neural networks~\cite{simonyan2014twostream, feichtenhofer2016twostreamfusion, wang2016tsn, zhu2017hidden, carreira2017i3d}, 3D convolutional networks~\cite{tran2018closer, ji2013conv3d, tran2015c3d, varol2018longterm, taylor2010convlearning, carreira2017i3d}, recurrent neural networks~\cite{baccouche2010soccer, ng2015beyond, donahua2017lrcn}, and pose-based methods~\cite{Yao, YAGG12, Gall, luvizon20182d3d, fani2017hockeyar, choutas2018potion}. \par
Besides the techniques mentioned above, pose features are widely used in works on action recognition. Pose estimation and action recognition are two problems that leverage information from each other. Yao \etal~\cite{Yao, YAGG12} claim that pose-level features are useful for action recognition and introduce an architecture for coupled 3D pose estimation and action recognition. Gall \etal~\cite{Gall} also use action recognition for 3D pose estimation. Luvizon \etal~\cite{luvizon20182d3d} use a multi-task framework for joint action recognition and 2D/3D pose estimation. Wang \etal~\cite{approach-pose-based-action-recognition-2} develop action representations based on 2D human poses. Fani \etal~\cite{fani2017hockeyar} use 2D pose from stacked hourglass network to infer action from still images. Iqbal \etal~\cite{Iqbal} introduce a framework to help estimate pose with action priors and then improve action priors with updated pose information and hence, oscillate between pose estimation and action recognition. Nie \etal~\cite{Nie} combine action recognition and video pose estimation in a unified framework with a spatial-temporal And-Or Graph model. Ch\'eron \etal~ \cite{cheronICCV15} use a pose-based CNN as a descriptor for action recognition.

Pose is a high-level spatial feature, while optical flows represent temporal information. Two-stream networks, ~\cite{feichtenhofer2016twostreamfusion, zhu2017hidden, girdhar2017vlad, carreira2017i3d, wang2018temporal, gao2018im2flow} is one of the prominent category of the state-of-the-art approaches in recent years, first proposed in~\cite{simonyan2014twostream}. A spatial stream analyzes a single video frame and a temporal stream uses multi-frame optical flow, both via a series of convolutions and fully-connected layers. Classification scores predicted by the two streams with softmax are fused via averaging or linear SVM. One of the advantage of separate streams is that they can be trained independently, and thereby spatial stream can be pre-trained on large still image classification datasets (such as ImageNet).

Plenty of variants of two-stream networks exist. Instead of fusing classification outputs of the two streams, Feichtenhofer \etal~\cite{feichtenhofer2016twostreamfusion} fuses the two streams at an intermediate convolutional layer with 3D convolutions, which are able to learn the correspondences between feature maps in two streams, into spatiotemporal feature maps. Spatiotemporal feature maps are also fused over time in order to consider a larger temporal scale. Wang \etal~\cite{wang2016tsn} model long-range temporal structure by uniformly segmenting videos and selecting a snippet from each segment. Two-stream networks are applied to each snippet and results are fused. Zhu \etal~\cite{zhu2017hidden} learn to estimate optical flow with an unsupervised architecture. It is done by minimizing the difference between the first frame and the frame reconstructed from the second frame by inverse warping according to predicted optical flow. Carreira \etal~\cite{carreira2017i3d}, use 3D convolutions in a two-stream architecture by pre-training original 2D filters on ImageNet and inflating them into 3D by repeating weights.

Our work is similar to these two-stream-based methods in the sense that we extract pose information and temporal information (optical flow parsed by CNN) in two separate streams before combining them. The pose stream is trained using transfer learning using pre-trained weights of the Part Affinity Fields model \cite{cao2017realtime} trained on MSCOCO dataset. 

In the context of hockey, tracking is a major research focus~\cite{cai2006robust, li2009video, okuma2013self, okuma2004boosted, musa2016modelbased, li2014hoglbp}. Most of the works on action recognition in hockey look at the game with a wider perspective, keeping event detection as the main focus~\cite{carbonneau2015playbreak, tora2017puck, sozykin2018imbalanced}. Action recognition, paying attention to individual players, is explored in very few works~\cite{lu2006arpcahog, lu2006ardistance, lu2009arboosted, fani2017hockeyar}. In these works, hand-crafted HOG features are first computed for tracking multiple individuals, and then a probabilistic framework is devised to model the action. They do not leverage high-level unique-to-human feature such as pose. In Fani \etal~\cite{fani2017hockeyar}, pose is considered but temporal information is neglected. Analyzing spatial and temporal information in two streams of CNNs is a powerful technique in understanding spatiotemporal structure, and pose features can provide valuable information for analyzing actions. In our work, two-stream-based architecture, combining pose and optical flow, applied to hockey action recognition concentrated on individual player, is presented.

\section{Methodology}\label{section:methodology}

\subsection{Overview}


The overall network architecture, as shown in Fig.~\ref{fig:overall_architecture}, illustrates the proposed approach of implementing a two-stream network incorporating pose estimation, via the model using part affinity fields (PAFs)~\cite{cao2017realtime}, and optical flow estimation, via LiteFlowNet~\cite{hui2018liteflownet}. The network takes a sequence of three images as an input, which is then used in the first stream by spatially transforming and cropping the image to a pixel size of 368$\times$368, centering the person and applying the pose estimation model, which is described in Section~\ref{section:pose}. Afterwards, the pose features are then concatenated in a latent feature vector layer (Section~\ref{section:latent}). The second stream then applies optical flow estimation to extract features at a macroscopic level (Section~\ref{section:recognition}). From both the streams, the action of the given sequence is then classified and the output of the network determines whether a hockey player is skating forward, skating backwards, passing or shooting. The training details are illustrated in Section~\ref{section:details}.

\subsection{Pose Estimation}\label{section:pose}

\begin{figure}[t]
\begin{center}
\includegraphics[width=0.98\linewidth]{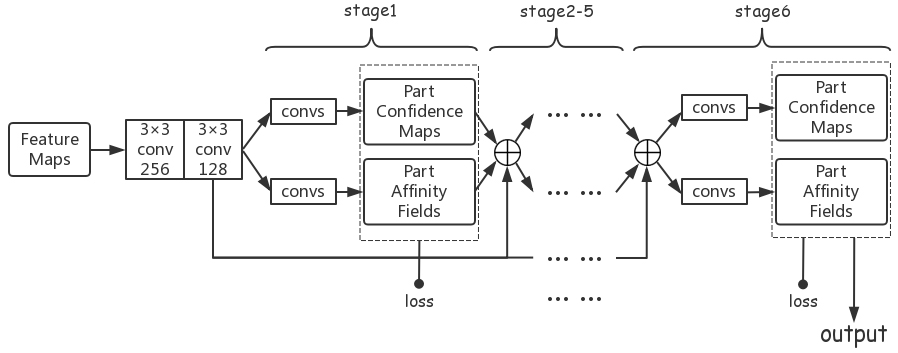}
\end{center}
   \caption{Multi-stage pose estimation architecture. Each stage predicts part confidence maps and part affinity fields through a series of convolutions. Prediction is iteratively refined and loss is computed at the end of each stage.}
\label{fig:pose_architecture}
\end{figure}

Cao \etal~\cite{cao2017realtime} propose a novel feature representation called part affinity fields, which evaluates association between two joints. In PAFs the, 2D vector at each pixel indicates position and orientation for a certain limb~\cite{cao2017realtime}. Fig.~\ref{fig:pose_architecture} shows the network generating part confidence maps and PAFs.

The feature maps extracted by VGG-19~\cite{Simonyan14c}, after two 3$\times$3 convolutions, are passed through six stages. Each stage is split into two branches predicting part confidence maps and part affinity fields via a series of convolutions. Then part confidence maps and part affinity fields as well as the aforementioned feature maps (passed through two convolutions) are concatenated together and taken as input by the next stage. Stage 1 has five convolutions, where the first three employ a kernel size of 3$\times$3 and the last two employ a kernel size of 1$\times$1. Stage 2-6 each has seven convolutions, where the first three employ a kernel size of 7$\times$7 and the last two employ a kernel size of 1$\times$1. Strides of all convolutions are 1, and paddings are all set to keep the size of the feature maps same. The prediction is refined iteratively, and loss is calculated for maps and fields output by every stage.

\subsection{Latent Feature Transformer}\label{section:latent}

\begin{figure}[t]
	\begin{center}
		\includegraphics[width=0.96\linewidth]{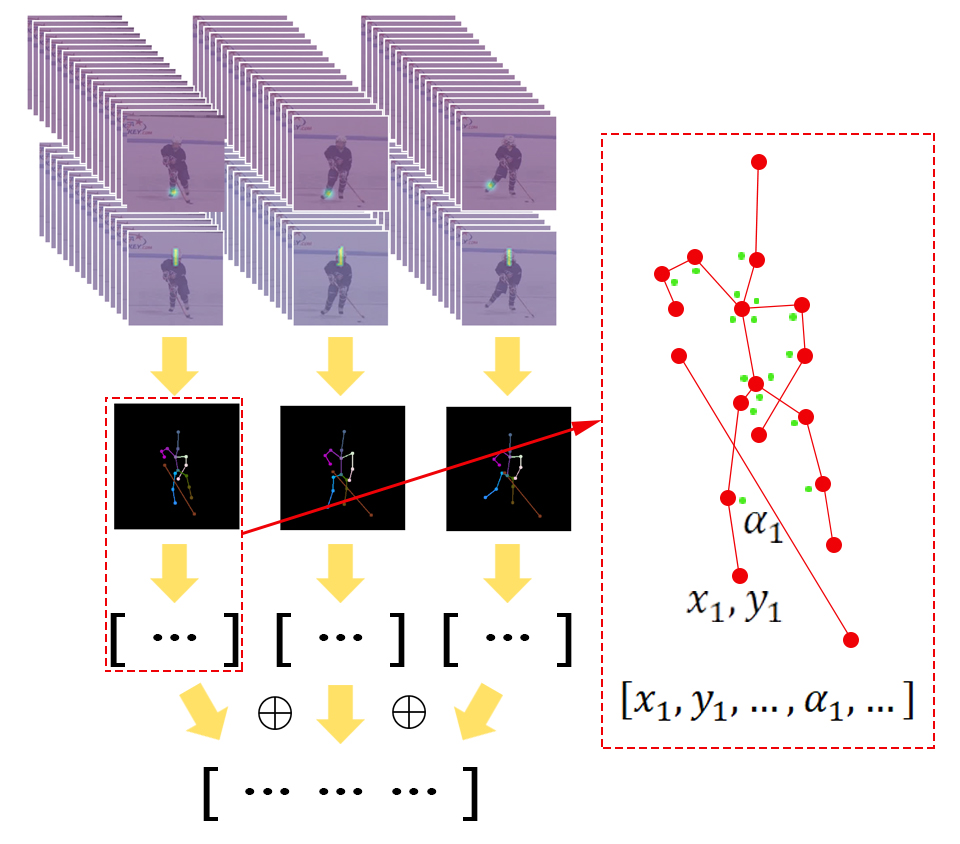}
	\end{center}
	\caption{Latent feature transformer. Pose is obtained from part confidence maps and part affinity fields for each image, and transformed into a flat latent joint feature vector. Dashed box on the right shows details of this transformation. The vector contains coordinates of all joints and angles between some limbs (green dots indicate angles between limbs that are to be calculated, which are shown more clearly in Table.~\ref{table:angles}). Finally, latent feature vectors of 3 images are concatenated.}
	\label{fig:latent_architecture}
\end{figure}

Fig.~\ref{fig:latent_architecture} briefly shows the pipeline for transforming part confidence maps and part affinity fields to a latent joint feature vector. To obtain the full pose of a single person, an existing algorithm~\cite{cao2017realtime} is modified, which first obtains limb connection candidates and then assembles them into pose of multiple persons. For each joint, we reserve two peaks with the highest score in corresponding part confidence map, instead of filtering candidates with threshold. This ensures
no joint will be lost. The joint with the highest score is not selected because the best location cannot be determined merely according to part confidence maps because the network sometimes makes mistakes, and that we want to leverage information provided by PAFs.

Then, a single candidate is selected for each joint. We start from the candidate of head top with the higher value, and expand it into full pose by iteratively selecting joint candidates which are most probable to associate with determined joints. Head top, being a relatively easier joint to detect as compared to limbs, is set to be the starting point since the network is less likely to make mistakes on it. The score of association between joint candidates is determined by calculating the line integral over the corresponding PAF along the limb, formally shown by Eq. (10) and (11) in Cao \etal~\cite{cao2017realtime}. Other joints that are easy to predict, such as the pelvis,  were tried as the starting point, however, the accuracy is nearly the same.

After locations of all joints in three images are obtained, the procedure mentioned in Fani \etal~\cite{fani2017hockeyar} is applied to each one of them. In Fani \etal~\cite{fani2017hockeyar}, joints identified in all images are scaled by the average head segment length (distance between head top and upper neck) in all training images. We normalize joints of each image with the head segment length in order to eliminate the impact of discrepancy in human's size between different images. Angles between certain limbs are also calculated (Table~\ref{table:angles}). Scaled joint locations and computed angles are concatenated to form a feature vector for each image. We concatenate vectors for three images into a one dimensional feature vector of size 156 which is fed to an action recognition component.

\begin{table}
    \centering
    \begin{tabular}{ccc}
        \hline
        head top & upper neck & thorax \\
        upper neck & thorax & left shoulder \\
        pelvis & thorax & left shoulder \\
        thorax & left shoulder & left elbow \\
        left shoulder & left elbow & left wrist \\
        upper neck & thorax & right shoulder \\
        pelvis & thorax & right shoulder \\
        thorax & right shoulder & right elbow \\
        right shoulder & right elbow & right wrist \\
        thorax & pelvis & left hip \\
        pelvis & left hip & left knee \\
        left hip & left knee & left ankle \\
        thorax & pelvis & right hip \\
        pelvis & right hip & right knee \\
        right hip & right knee & right ankle \\
        left hip & pelvis & right hip \\
        \hline
    \end{tabular}
    \caption{Angles calculated in latent feature transformer. Each row in the table indicates an angle. A row whose items are $A$, $B$, $C$ from left to right represents $\angle ABC$.}
    \label{table:angles}
\end{table}

\subsection{Action Recognition Component}\label{section:recognition}

LiteFlowNet (Hui \etal~\cite{hui2018liteflownet}) is a state-of-the-art network for optical flow estimation. In their work, pyramidal features are received by cascaded flow inference and flow regularization modules, which iteratively increase resolution of flow fields. Pre-trained LiteFlowNet is used in our pipeline. Since LiteFlowNet takes two images as input, two optical flows are generated from three images.

\begin{figure}[t]
\begin{center}
\includegraphics[width=0.7\linewidth]{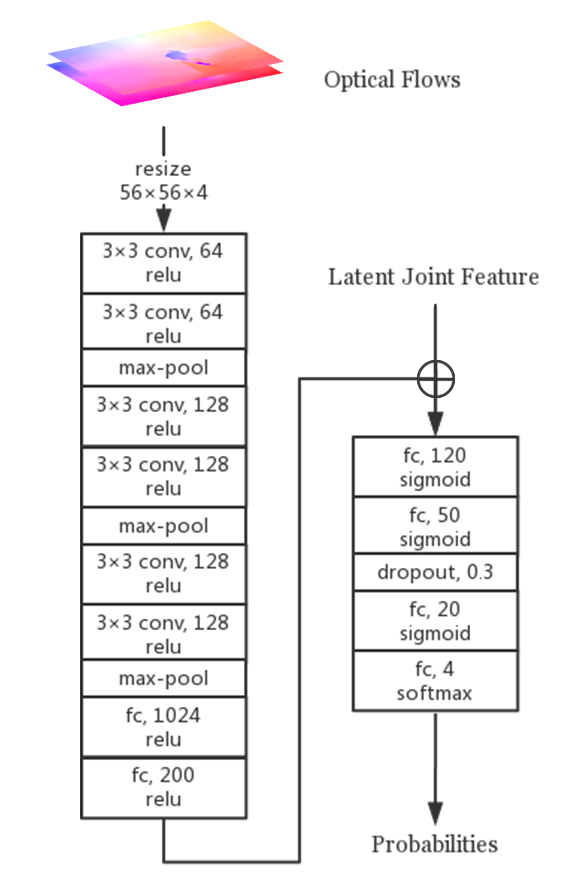}
\end{center}
   \caption{Action recognition architecture. Optical flows are resized and passed through interwoven convolutions and max-pooling, followed by fully-connected layers, and converted into a vector. It is concatenated with latent joint feature vector. Fully-connected network predicts probabilities of each class from the vector.}
\label{fig:action_architecture}
\end{figure}

The action recognition component leverages information provided by joint locations and optical flows. The architecture is illustrated in Fig.~\ref{fig:action_architecture}. The optical flow fields obtained are concatenated into a 4-channel map and resized to 56$\times$56 pixels. Then, the map is passed through several convolutional and max-pooling layers followed by two fully-connected layers and converted into a flat feature vector. Relu activation is used for all convolutional and fully-connected layers in this part. The feature vector generated from optical flows is concatenated with the latent joint feature vector. The flow feature vector concatenated with latent joint feature vector is passed through four fully-connected layers, the first three of them with sigmoid activation and the last with softmax to output probabilities of four classes. A dropout layer is added after the second fully-connected layer (50 units) to reduce overfitting.

\subsection{Training Details}\label{section:details}

Since our dataset is small, various methods are employed to reduce overfitting such as dropout, dataset augmentation and early stopping.

As a basic method of transfer learning, pose estimation network is fine-tuned based on weights pre-trained on MSCOCO dataset \cite{coco}. The dataset contains over 100K person instances and covers various real-world scenarios, which can provide relevant knowledge for transfer learning. In order to avoid overfitting, weights of all layers, except last 3 layers of the stage 5 and all layers of the stage 6 are frozen. However, joints we want to learn here are different from that in the MSCOCO dataset. In order to perform transfer learning, only the last two stages of the pose estimation network are trained such that, the loss of the rest of the stages is not computed and the last two stages output 18 new joints.

A variety of data augmentation is performed in training to make our dataset appear more diverse. For the pose estimation network, original images are randomly flipped, scaled, rotated, similar to Cao \etal~\cite{cao2017realtime}. For the action recognition network, in addition to the methods applied to pose estimation network, we perturb the location of each joint.
Note that whenever flipping is applied to joints while training action recognition component, it is also applied to optical flows, because direction of background movement and orientation of person, which is represented by pose, are tied together when telling the direction of person's movement.


The pose estimation network and the action recognition network are trained separately. Validation loss of the pose estimation network decreases with training loss, so the network is trained until convergence and the last model checkpoint is selected. However, the action recognition part starts to overfit after 30 epochs. So an early stopping technique of training 30 epochs and picking up the checkpoint with highest validation accuracy was adopted. We select three models with highest validation accuracy at test time, which will be explained in Section~\ref{section:test_ar}.

In addition, we found that the action recognition network is difficult to train if the input is the prediction of the pose estimation network, so we instead train the pose estimation network with augmented ground truth of joint locations and validate it with the prediction. The network generalizes well to the case where joints are not precisely localized. This is because augmentation applied to joints eliminates the impact of possible discrepancy between distribution of ground-truth and predicted joint locations, which makes the network able to tolerate joint errors.

The training hyperparameter configurations are as follows. Weights are learned using mini-batch stochastic gradient descent with batch size set to 2 and momentum set to 0.9 for both two sub-networks. For the pose estimation network, L2 regularization is added to the convolution kernel weights with a regularization coefficient of $5\times10^{-4}$. In every epoch, all training images are fed once, so the number of iterations per epoch is $\frac{N}{2}$ (N training images). Training lasts 300 epochs. The learning rate is initially set to $10^{-2}$ and changed to $10^{-3}$ after 200 epochs. For the action recognition network, learning rate is $10^{-2}$ throughout the 30-epoch training. The dropout ratio is set to 0.3. The training process takes about 14 hours for the pose estimation network and about 13 minutes for the action recognition network, on a TITAN X GPU.

\section{Testing and Results}\label{section:test}

\subsection{Dataset Preparation}

The model is trained and tested on the HARPET dataset which is composed of sequences of 3 images with time interval of $\frac{1}{6}$ seconds between any two successive frames in 30 frames per second video. Sequence length is set according to previous work on temporal modeling~\cite{gaidon2013temporal, wang2014latent, wang2016tsn}. The sequences are collected from video clips scraped from the internet and from several instructional DVDs about hockey. Video segments were extracted from these videos and broken up into consecutive frames.  Next, the sequences are classified into one of 4 classes: forward, backward, passing and shooting. Finally, 18 joints (16 human joints and 2 stick joints) are annotated in all images.

The dataset has 106 sequences for forward, 104 for backward, 113 for passing and 101 for shooting. There are a total of 1272 images of which joints are annotated respectively. The HARPET dataset is randomly split into three sets: 70\% for training, 15\% for validation and 15\% for testing. The pose estimation component and the action recognition component are both trained on training set. The validation set is used to pick the best model.


\subsection{Accuracy of Pose Estimation}

To evaluate the pose estimation network, PCKh@0.5~\cite{mpii} metric is used. According to the PCKh@0.5 metric, a joint is localized correctly if distance between prediction and ground truth is less than one-half of head segment length (distance between top of head and upper neck), and percentage of correctly-localized joints is computed. Results are illustrated in Table.~\ref{table:pose}.

The results demonstrates that the network trained on MSCOCO dataset can be transferred to the hockey domain with good accuracy (86.95\% overall accuracy). Stick prediction has the worst precision (75.40\%), which, several reasons account for poor precision of the hockey stick. (1) Joints, minus the hockey stick, are inferred from joints considered in the MSCOCO dataset which makes it easier for the network to transfer those joints, however, the stick is a new concept which takes more effort to learn. (2) The current model does not have a large enough receptive field to capture the whole stick that can be very long in images. (3) In many images, the stick is occluded or moves too quickly, adding difficulties to recognition. Prediction of elbows and wrists is also unsatisfactory, due to frequent occlusions.

\begin{table}
    \centering
    \begin{tabular}{c|c}
        \hline
        Parts & PCKh@0.5 (left/right, top/end) \\
        \hline
        Head & 94.18\% \\
        Upper Neck & 97.25\% \\
        Thorax & 96.30\% \\
        Shoulder & 85.19\%/89.42\% \\
        Elbow & 78.31\%/80.95\% \\
        Wrist & 76.72\%/80.42\% \\
        Pelvis & {\bf 97.35\%} \\
        Hip & 92.06\%/87.83\% \\
        Knee & 91.53\%/91.00\% \\
        Ankle & 89.42\%/86.24\% \\
        Stick & 71.96\%/78.84\% \\
        Overall & 86.95\%
    \end{tabular}
    \caption{Results of pose estimation. Values of left and right shoulder/elbow/wrist/hip/knee/ankle, as well as stick top and stick end, are averaged to shorten the table.}
    \label{table:pose}
\end{table}

\begin{figure}[t]
	\begin{center}\hfill
		\subfloat{\includegraphics[width=0.38\linewidth]{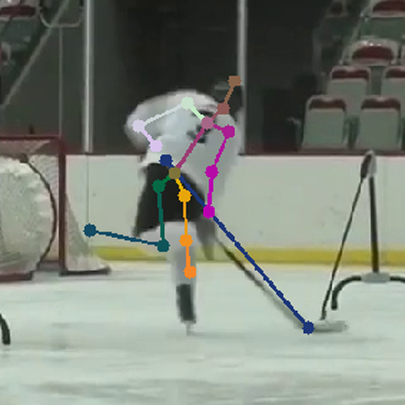}} \hfill
		\subfloat{\includegraphics[width=0.38\linewidth]{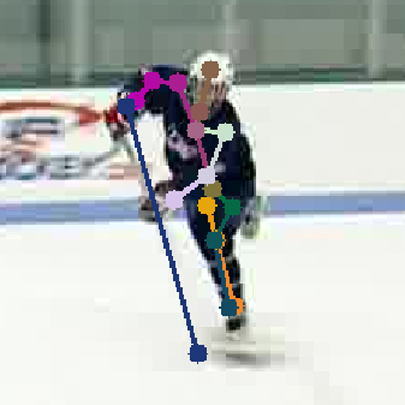}} \hfill \\ \hfill
		\subfloat{\includegraphics[width=0.38\linewidth]{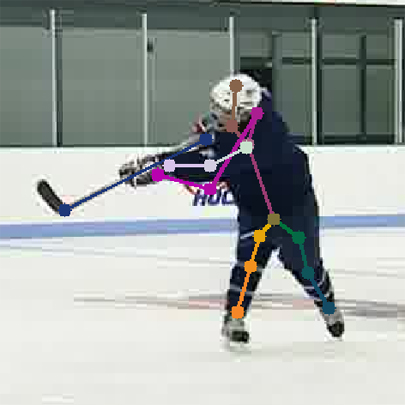}} \hfill
		\subfloat{\includegraphics[width=0.38\linewidth]{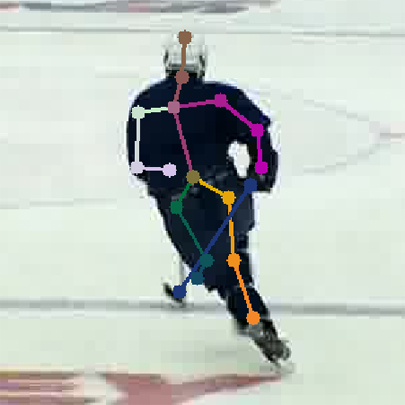}} \hfill
	\end{center}
	\caption{Common failure cases of pose estimation.}
	\label{fig:pm_failures}
\end{figure}

Common failure cases are shown in Fig.~\ref{fig:pm_failures}. Left-and-right error and stick mislocalization due to occlusion and high-speed motion are typical.

\begin{figure*}
	\begin{center}
		\subfloat[Passing]{
		    \includegraphics[width=0.16\linewidth]{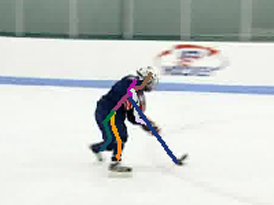}
		    \includegraphics[width=0.16\linewidth]{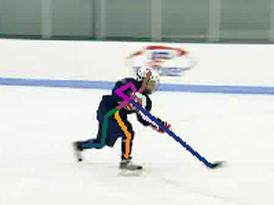}
		    \includegraphics[width=0.16\linewidth]{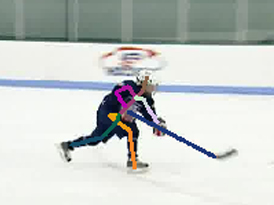}
		}
		\subfloat[Passing]{
		    \includegraphics[width=0.16\linewidth]{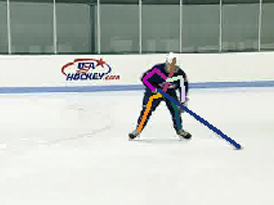}
		    \includegraphics[width=0.16\linewidth]{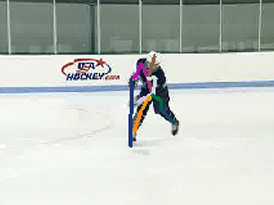}
		    \includegraphics[width=0.16\linewidth]{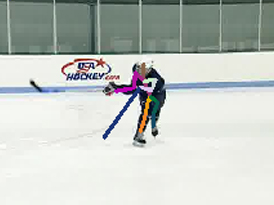}
		} \\
		\subfloat[Backward]{
		    \includegraphics[width=0.16\linewidth]{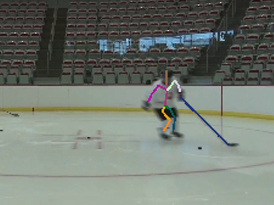}
		    \includegraphics[width=0.16\linewidth]{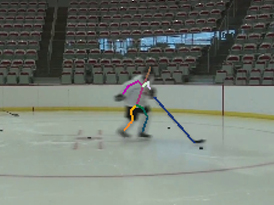}
		    \includegraphics[width=0.16\linewidth]{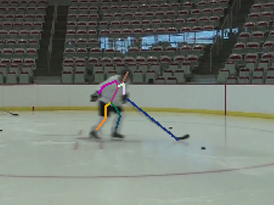}
		}
		\subfloat[Shooting]{
		    \includegraphics[width=0.16\linewidth]{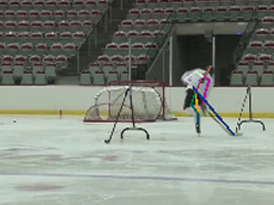}
		    \includegraphics[width=0.16\linewidth]{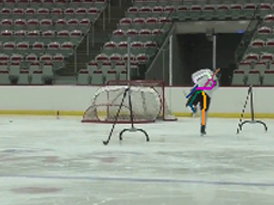}
		    \includegraphics[width=0.16\linewidth]{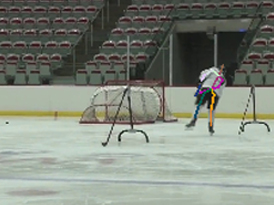}
		}
	\end{center}
	\caption{Some examples of correct classification. Action recognition network can tolerate joint localization errors.}
	\label{fig:ar_correct}
\end{figure*}

\begin{figure*}
	\begin{center}
		\subfloat[Forward$\rightarrow$Passing]{
		    \includegraphics[width=0.16\linewidth]{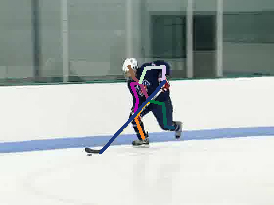}
		    \includegraphics[width=0.16\linewidth]{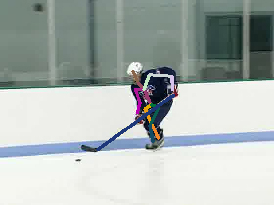}
		    \includegraphics[width=0.16\linewidth]{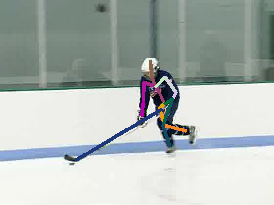}
		}
		\subfloat[Forward$\rightarrow$Shooting]{
		    \includegraphics[width=0.16\linewidth]{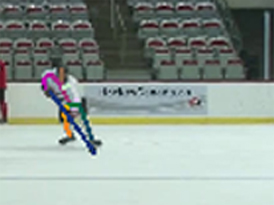}
		    \includegraphics[width=0.16\linewidth]{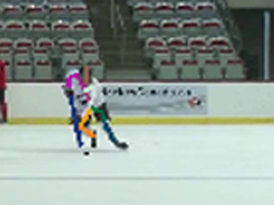}
		    \includegraphics[width=0.16\linewidth]{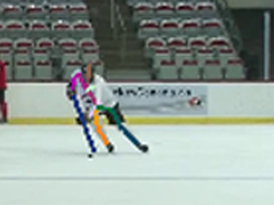}
		} \\
		\subfloat[Shooting$\rightarrow$Forward]{
		    \includegraphics[width=0.16\linewidth]{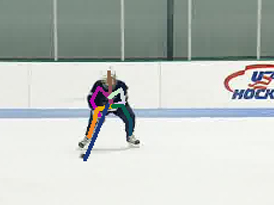}
		    \includegraphics[width=0.16\linewidth]{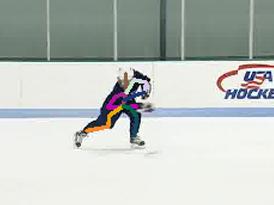}
		    \includegraphics[width=0.16\linewidth]{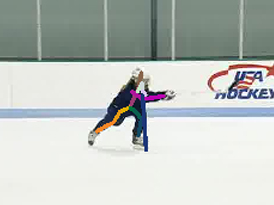}
		}
		\subfloat[Shooting$\rightarrow$Passing]{
		    \includegraphics[width=0.16\linewidth]{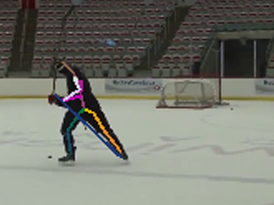}
		    \includegraphics[width=0.16\linewidth]{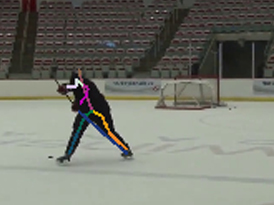}
		    \includegraphics[width=0.16\linewidth]{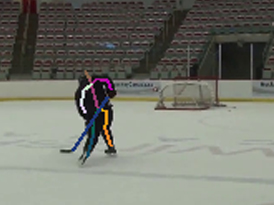}
		}
	\end{center}
	\caption{Common failure cases of action recognition (ground truth$\rightarrow$prediction).}
	\label{fig:ar_failures}
\end{figure*}

\begin{table}
    \centering
    \begin{tabular}{c|cccc}
        \hline
        \multirow{2}{*}{Methods} & \multicolumn{4}{c}{Precision} \\
        & Fw. & Bw. & Ps. & St. \\
        \hline
        -ST, -OF & 69.25\% & 81.82\% & 46.18\% & 82.63\% \\
        -ST, +OF & 77.73\% & 82.05\% & 49.26\% & 77.22\% \\
        +ST, -OF & 79.12\% & 88.08\% & 53.36\% & 80.65\% \\
        +ST, +OF & {\bf 96.06\%} & {\bf 95.83\%} & {\bf 56.67\%} & {\bf 87.22\%} \\
        \hline
        \hline
        \multirow{2}{*}{Methods} & \multicolumn{4}{c}{Recall} \\
        & Fw. & Bw. & Ps. & St. \\
        \hline
        -ST, -OF & 71.67\% & 50.00\% & 66.67\% & 80.39\% \\
        -ST, +OF & 70.00\% & 50.00\% & 73.33\% & {\bf 88.24}\% \\
        +ST, -OF & {\bf 80.00\%} & 62.50\% & 83.33\% & 72.55\% \\
        +ST, +OF & {\bf 80.00\%} & {\bf 87.50\%} & {\bf 90.00\%} & 80.40\% \\
        \hline
    \end{tabular}
    \caption{Precision and recall rate of each combination, for 4 classes. Each value in the table is the average of corresponding values of 3 checkpoints.}
    \label{table:pr}
\end{table}

\begin{table}
    \centering
    \begin{tabular}{c|cccc}
        \hline
        \multirow{2}{*}{Methods} & \multicolumn{4}{c}{Accuracy} \\
        & $1^{st}$ & $2^{nd}$ & $3^{rd}$ & Avg. \\
        \hline
        -ST, -OF & 71.43\% & 68.25\% & 63.49\% & 67.72\% \\
        -ST, +OF & 68.25\% & 68.25\% & 74.60\% & 70.37\% \\
        +ST, -OF & 74.60\% & 73.02\% & 74.60\% & 74.07\% \\
        +ST, +OF & 84.13\% & {\bf 85.71\%} & 80.95\% & {\bf 83.60\%} \\
        \hline
    \end{tabular}
    \caption{Overall accuracy of each combination. Results of all selected checkpoints are shown as well as average values. ${\bm 1^{st}}$, ${\bm 2^{nd}}$, ${\bm 3^{rd}}$ refer to validation accuracy ranking}
    \label{table:acc}
\end{table}

\begin{figure}[t]
	\begin{center}
		\subfloat[{\bf -ST, -OF}]{\includegraphics[width=0.48\linewidth]{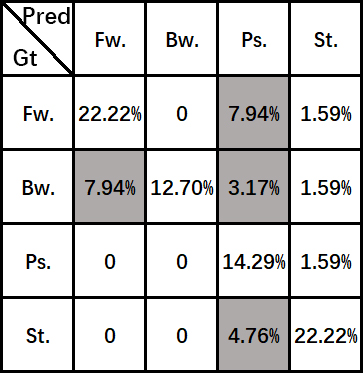}}\hfill
		\subfloat[{\bf -ST, +OF}]{\includegraphics[width=0.48\linewidth]{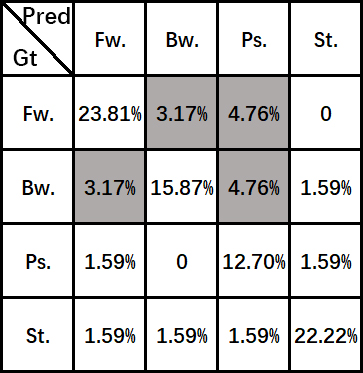}} \\
		\subfloat[{\bf +ST, -OF}]{\includegraphics[width=0.48\linewidth]{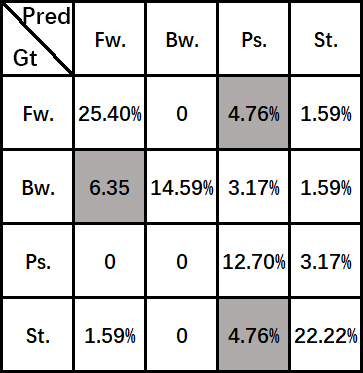}}\hfill
		\subfloat[{\bf +ST, +OF}]{\includegraphics[width=0.48\linewidth]{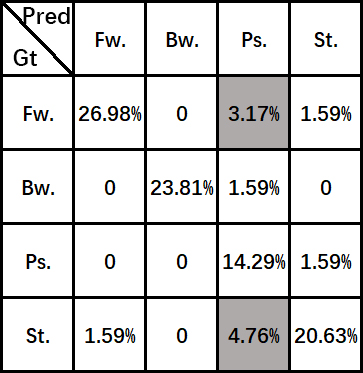}}
	\end{center}
	\caption{Confusion matrices for 4 combinations. In each cell is percentage of sequences which belong to a certain class and are mistaken for a certain class. Cells indicating misclassfication with ratio higher than 3\% are highlighted.}
	\label{fig:confusion}
\end{figure}

\subsection{Accuracy of Action Recognition}\label{section:test_ar}




 To show that the hand-held object is a strong cue for action recognition in hockey, coordinates of stick top (butt end) and stick end (stick blade) are purposely ignored by latent feature transformer (denoted by {\bf -ST}) for a comparison with the original method which takes all joints including the stick into consideration ({denoted by \bf +ST}). Besides, the temporal stream which analyzes optical flows is removed so that the action recognition network only looks at the latent joint feature vector (denoted by {\bf -OF}), for a comparison with the original two-stream architecture (denoted by {\bf +OF}). Hence, we have four combinations.

As mentioned above, since the validation accuracy does not steadily increase throughout training, selecting a checkpoint with the highest validation score is appropriate compared to simply picking up the last checkpoint. Moreover, due to lack of data, validation accuracy fluctuates drastically throughout training and there is an inevitable gap between validation and test accuracies. To evaluate the overall performance of each combination more appropriately, we test on 3 checkpoints of action recognition model with top 3 validation accuracy.

Table~\ref{table:pr} shows the precision and recall for each class of each combination ({\bf Fw.}=Forward, {\bf Bw.}=Backward, {\bf Ps.}=Passing, {\bf St.}=Shooting for convenience). Here values of 3 checkpoints of every combination is averaged. Overall accuracy of each combination is illustrated in Table~\ref{table:acc}. In this table, the results of all selected checkpoints are shown (${\bm 1^{st}}$, ${\bm 2^{nd}}$, ${\bm 3^{rd}}$ refer to validation accuracy ranking) along with average values.

From the macro view, both stick information and optical flow, complementary to each other, help improve the overall accuracy as well as precision and recall rate for most classes. Exceptions lie in precision and recall rate of shooting and recall rate of skating forward. The stick plays a more important role than optical flow, as indicated by the combination of {\bf +ST, -OF} outperforms {\bf -ST, +OF}. When leveraging both stick and optical flow, end-to-end action recognition accuracy can be boosted to about 85\%.

From the micro view, precision rate for passing is unsatisfying while the recall rate for passing is comparable to other classes, which means in many cases, other actions are mistaken for passing. This can also be seen from confusion matrices (Fig.~\ref{fig:confusion}). From comparison between {\bf -ST, -OF} and {\bf +ST, -OF} as well as {\bf -ST, +OF} and {\bf +ST, +OF}, stick increases accuracy of other 3 classes except shooting. This can be justified by observing that shooting is the only action among the 4 that is likely to lead to drastic change in pose so that joints are sufficient to recognize it. From comparison between {\bf -ST, -OF} and {\bf -ST, +OF} as well as {\bf +ST, -OF} and {\bf +ST, +OF}, it is demonstrated that optical flows improve results under most circumstances.

Some examples of correct classification are displayed in Fig.~\ref{fig:ar_correct}. In Fig.~\ref{fig:ar_correct} (a) and (c), pose is correctly obtained thus leading to correct classification. In contrast, Fig.~\ref{fig:ar_correct} (b) and (d), joints (sticks, more specifically) failed to be localized accurately but still produced correct results in the action recognition network, thus indicating that the action recognition network can tolerate some joint localization errors.

Fig.~\ref{fig:ar_failures} shows some failure cases. It can be seen that accuracy of action recognition is limited by accuracy of pose estimation. Misclassification of Fig.~\ref{fig:ar_failures} (c) and (d) is due to the failure in predicting stick top and stick end. This is common in shooting case because the stick is likely to move too fast, or be lifted too high (lifting stick too high is a rare case in training images, so the network cannot recognize the stick well in this situation). Fig.~\ref{fig:ar_failures} (a) and (b) reveal a major inherent downside of the method. Even if pose is predicted precisely, correctness cannot be guaranteed. Under many circumstances, contextual information is helpful, such as movement of the puck, position of the goal, action of surrounding players. Pose does not contain these factors, and it is also difficult for the network to learn to capture crucial detailed information from optical flows, especially when the dataset is too limited.

\section{Conclusion}


In this paper, we propose a novel two-stream architecture for action recognition. The two streams estimate pose and parse optical flows via CNN, which are then concatenated and passed through fully-connected layers to output classification scores. The architecture extends general two-stream networks by leveraging pose, which is a high-level feature that is shown to be suitable for action recognition, achieving 85\% end-to-end accuracy. Experimental results demonstrate that pose and optical flows, as different-level features, are complementary to each other. It is also demonstrated that hand-held objects, sticks in hockey context, play an important role in analyzing the sport actions. In addition, we transfer the information from the pose estimation model pre-trained on MSCOCO dataset to our small hockey dataset achieving 87\% overall accuracy measured by PCKh@0.5.

There is room for improvement. (1) Although three sparsely-sampled images are generally adequate to depict an action, considering additional images can be more reliable and accurate. (2) Sometimes, a joint in an image that is difficult, even for a human to localize, can be better inferred by utilizing the neighboring frames i.e, temporal information can also be leveraged in pose estimation. (3) High-level activities such as puck location and goal scored, can also be taken into consideration.

{\small
\bibliographystyle{ieee}
\bibliography{ref}
}

\end{document}